\documentclass{article}
\usepackage{spconf,amsmath,epsfig}
\usepackage{amsfonts}
\usepackage{xcolor}
\usepackage{subfigure}
\usepackage{multirow}

\title{SA-Modified: A Foundation Model-Based Zero-Shot Approach for Refining Noisy Land-Use Land-Cover Maps}
\name{Sparsh Pekhale, Rakshith Sathish, Sathisha Basavaraju, Divya Sharma}
\address{SatSure Analytics Pvt. Ltd., Bengaluru, India}
\begin{document}
%
\maketitle
\begin{abstract}
Land-use and land cover (LULC) analysis is critical in remote sensing, with wide-ranging applications across diverse fields such as agriculture, utilities, and urban planning. However, automating LULC map generation using machine learning is rendered challenging due to noisy labels. Typically, the ground truths (e.g. ESRI LULC, MapBioMass) have noisy labels that hamper the model's ability to learn to accurately classify the pixels. Further, these erroneous labels can significantly distort the performance metrics of a model, leading to misleading evaluations. Traditionally, the ambiguous labels are rectified using unsupervised algorithms. These algorithms struggle not only with scalability but also with generalization across different geographies. To overcome these challenges, we propose a zero-shot approach using the foundation model, Segment Anything Model (SAM), to automatically delineate different land parcels/regions and leverage them to relabel the unsure pixels by using the local label statistics within each detected region. We achieve a significant reduction in label noise and an improvement in the performance of the downstream segmentation model by $\approx 5\%$ when trained with denoised labels.

\end{abstract}
\section{Introduction}
\label{sec:intro}

\begin{figure}[!h]
\centering
\subfigure[Noisy Ground Truth]{\includegraphics[width=0.7\linewidth]{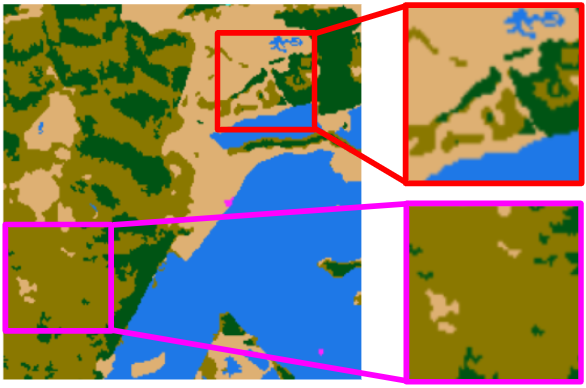}\label{fig:gt}}
\subfigure[Denoised Ground Truth]{ \includegraphics[width=0.7\linewidth]{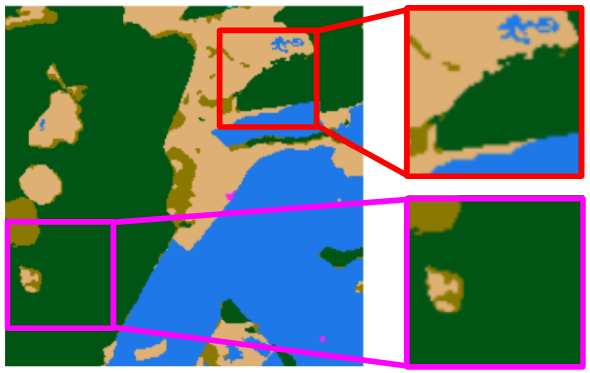}}
\caption{\textbf{Graphical Abstract}. (a) depicts the noisy ground truth annotations, which contain incorrect and ambiguous labels. (b) shows the denoised ground truth after applying the proposed zero-shot approach using the Segment Anything Model (SAM), resulting in cleaner and more reliable labels. The zoomed-in regions highlight the improvements in label accuracy achieved by the method.}
\label{fig:abstract}
\end{figure}

Accurate Land-Use Land Cover (LULC) mapping is essential for numerous remote sensing applications, including crop monitoring, urban infrastructure development, and environmental conservation. To generate LULC maps at scale, we often rely on supervised machine learning models. However, this automation faces significant hurdles, primarily stemming from the quality and consistency of ground truth annotations. In many widely used LULC datasets, such as ESRI LULC \cite{esri} and MapBioMass \cite{mapbiomas}, the presence of noisy, incorrect, or ambiguous labels is a persistent issue. These inaccuracies not only degrade model performance but also undermine the reliability of subsequent analyses based on these maps.

\begin{figure*}[t]
    \centering
    \includegraphics[width=0.64\linewidth]{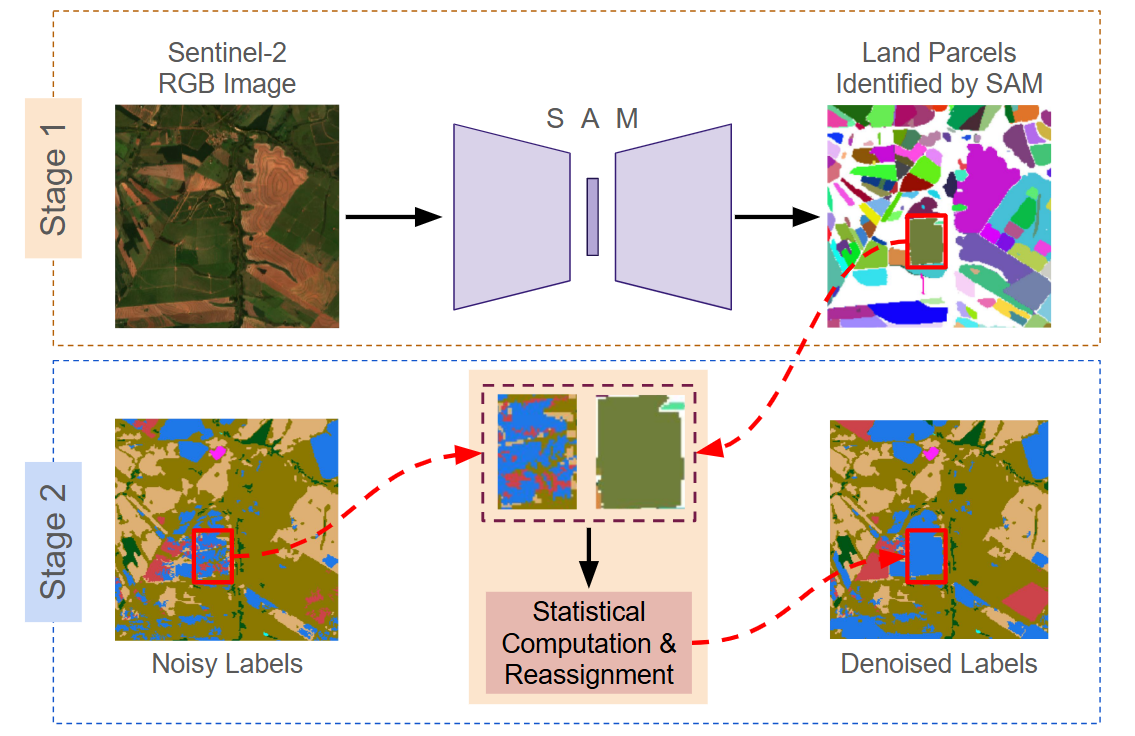}
    \caption{ \textbf{Overview of the Proposed Two-Stage Approach}. In \textit{Stage 1}, the Segment Anything Model (SAM) is employed to delineate distinct land parcels in the input imagery using zero-shot learning. \textit{Stage 2} involves analyzing the local label statistics within each identified parcel and reassigning labels based on the dominant class within each region.}
    \label{fig:method}
\end{figure*}

The challenges associated with noisy labels in LULC datasets are multi-faceted. First, the complexity of landscapes and the subtle differences between land cover classes can lead to misclassification during the model-based annotation process. This is especially problematic in heterogeneous environments where distinct land covers may share similar spectral signatures, making it difficult for automated systems to distinguish between them. While these datasets are generated through sophisticated algorithms, the inherent limitations of these models, including biases in training data and algorithmic assumptions, contribute to the propagation of errors. 

\noindent \textbf{Prior Arts:} Traditional approaches to addressing label noise, such as unsupervised clustering algorithms \cite{kmeans, dbscan}, attempt to segment the landscape into homogeneous parcels based on spectral and spatial features. While these methods can identify some discrepancies, they are limited by their dependence on predefined assumptions about the data's structure, which often fails to generalize across different regions. Moreover, these algorithms struggle with scalability, making them impractical for large-scale, global LULC mapping initiatives where consistency and adaptability are crucial.
To address these challenges, we propose a zero-shot approach \cite{zeroshot} utilizing the Segment Anything Model (SAM) \cite{sam}, a foundation model as an effective alternative to unsupervised algorithms.

\noindent \textbf{Our Approach}:
Our proposed method is structured as a two-stage approach designed to mitigate label noise in LULC datasets. In the first stage, we employ the Segment Anything Model (SAM), a foundation model utilizing zero-shot learning, to identify and delineate distinct land parcels within the input imagery. SAM's segmentation outputs are used to define these parcels, enabling us to capture regions with similar characteristics accurately. In the second stage, we analyze the local label statistics within each identified parcel. Specifically, we determine the majority class within each region and use this information to relabel ambiguous or noisy pixels. This process effectively reduces label noise by ensuring that pixel labels within each parcel align with the dominant class, leading to more accurate and reliable LULC maps.

\vspace{-0.35cm}
\section{Method}
\label{sec:method}

\noindent\textbf{Stage 1: SAM-based Land Parcel Identification}

In the first stage, we utilize the Segment Anything Model (SAM), a foundation model that combines deep convolutional neural networks (CNNs) and transformer architectures for state-of-the-art segmentation. Trained on a vast dataset, SAM excels in recognizing and delineating objects and regions across diverse and complex environments by generating a comprehensive feature map that captures both spatial details and contextual information. Unlike traditional models, SAM generalizes effectively without fine-tuning, making it ideal for segmenting input images into distinct land parcels.

Mathematically, let \( I \) represent the input image, and \( F(I) \) denote the feature map generated by SAM. For each pixel \( p_j \) in the image, SAM predicts the segment \( s_i \) it belongs to based on the feature representation \( F(p_j) \). This can be expressed as:

\[
s_i = \arg\max_{s \in S} P(s \mid F(p_j))
\]

where \( S \) is the set of all possible segments, and \( P(s \mid F(p_j)) \) is the probability that pixel \( p_j \) belongs to segment \( s \) given its feature representation \( F(p_j) \).

\noindent\textbf{Stage 2: Majority Voting for Class Refinement}

Once the land parcels have been identified in Stage 1, the second stage involves refining the noisy labels within each segment using a majority voting mechanism. For each segment \( s_i \) identified in Stage 1, we have a set of pixels \( P_i = \{p_1, p_2, \dots, p_{m_i}\} \) and their corresponding noisy labels \( L_i = \{l_1, l_2, \dots, l_{m_i}\} \). The refined label \( L_i' \) for each segment is determined by applying majority voting on the noisy labels within that segment:

\[
L_i' = \text{mode}(L_i) = \arg\max_{l} \sum_{j=1}^{m_i} \mathbb{I}(l_j = l)
\]

where \( \mathbb{I}(l_j = l) \) is an indicator function that equals 1 if the label \( l_j \) of pixel \( p_j \) matches the class \( l \), and 0 otherwise.

\section{Experiments}
\label{sec:experiments}

\textbf{Dataset:} 
The experiments are conducted using the MapBiomas LULC dataset for Brazil. This dataset provides annual LULC classifications in $30 m$ spatial resolution across Brazil. For our study, we focus exclusively on the level $1$ labels, which include Cropland, Forest, Barren/Built-up, Waterbody, and Pasture. Additionally, the dataset contains a class labelled "mosaic of uses," representing pixels where the classification is uncertain, and the exact land cover type is ambiguous. Our objective is to denoise these uncertain labels by identifying the most probable class to which these pixels belong. We use Harmonized Landsat and Sentinel-2 (HLS) as the satellite image source. A multi-level stratified sampling strategy was applied across Brazil to identify $112,092$ AOIs of $235.93 KM^2$ each for our experiments.

\noindent\textbf{\textit{Baseline 1 (BL1):}} K-means algorithm aims to minimize the variance within clusters by iteratively assigning data points to one of \( K \) clusters based on the nearest mean, or centroid, of the cluster. Mathematically, the objective of K-means is to minimize the following loss function:

\[
J = \sum_{k=1}^{K} \sum_{i=1}^{n_k} \| x_i^{(k)} - \mu_k \|^2
\]

where \( \mu_k \) is the centroid of the \( k \)-th cluster, \( x_i^{(k)} \) represents the \( i \)-th data point assigned to the \( k \)-th cluster, and \( n_k \) is the number of data points in cluster \( k \).

\noindent\textbf{\textit{Baseline 2 (BL2):}} Density-Based Spatial Clustering of Applications with Noise (DBSCAN) clusters data based on density. It requires two parameters: \( \epsilon \), the neighborhood radius, and  \( MinPts \), the minimum number of points to form a cluster.  Mathematically, a point \( p \) is a core point if:
\[
|\{ q \in D \mid \text{dist}(p, q) \leq \epsilon \}| \geq MinPts
\]
where \( D \) represents the dataset, and \( \text{dist}(p, q) \) is the distance between points \( p \) and \( q \). Clusters are formed by expanding from core points, while points that don't meet the criteria are labeled as noise.

\section{Results and Discussion}
\label{sec:results}
The reliability of the model is evaluated on four fronts; (I) the Ability to reduce the noise present in the form of stray pixels, (ii) the ability to improve boundary/demarcation between classes hence reducing the mixing between classes, (iii) improvement in downstream segmentation task and (iv) and comparison with other unsupervised algorithms

\noindent \textbf{A. Impact on Stray Pixels}

\begin{figure}[h]
\centering
\subfigure[HLS Input]{\includegraphics[width=0.3\linewidth]{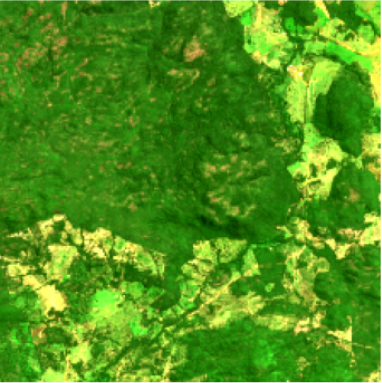}\label{fig:gt}}
\subfigure[Noisy GT]{ \includegraphics[width=0.3\linewidth]{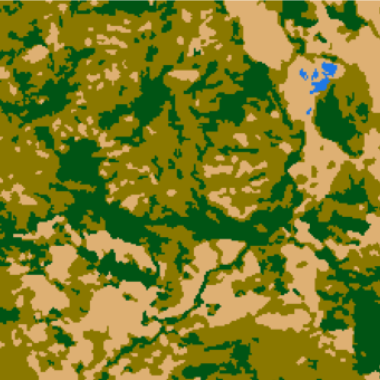}}
\subfigure[Denoised GT]{ \includegraphics[width=0.3\linewidth]{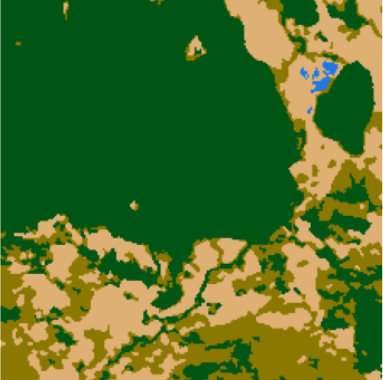}}
\caption{Reduction of label noise in stray pixels. (a) HLS input image, (b) Noisy ground truth with uncertain pixels labelled as 'mosaic of uses' (yellow), and (c) Denoised ground truth with these pixels reassigned to the forest class (dark green) based on SAM's segmentation.}
\label{fig:qa_stray}
\end{figure}

Fig. \ref{fig:qa_stray} shows the reduction in label noisy stray pixels. Most of the unsure pixels labelled as 'mosaic of uses' shown in yellow have been reassigned to forest class (shown in dark green), as the entire region was identified to belong to a single land segment/parcel by SAM.

\noindent \textbf{B. Impact on Class Boundaries}

\begin{figure}[h]
\centering
\subfigure[HLS Input]{\includegraphics[width=0.3\linewidth]{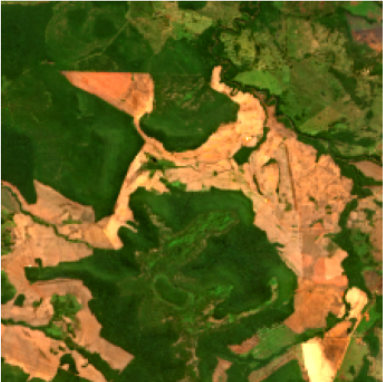}\label{fig:gt}}
\subfigure[Noisy GT]{ \includegraphics[width=0.3\linewidth]{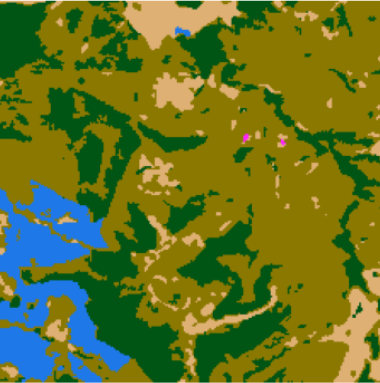}}
\subfigure[Denoised GT]{ \includegraphics[width=0.3\linewidth]{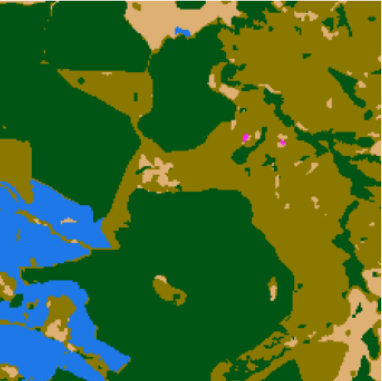}}
\caption{Improvements in class boundaries. (a) HLS input image, (b) Noisy ground truth with uncertain pixels (yellow), and (c) Denoised ground truth with more accurate boundaries between classes based on SAM's segmentation. }
\label{fig:qa_boundary}
\end{figure}

Fig. \ref{fig:qa_boundary} shows the impact of denoising on class boundaries. The boundaries between forest class and patches of cultivated/deforested land is observed. Further, a significant reduction in the mixing of class labels is also observed.

\begin{figure*}[h]
    \centering
    \includegraphics[width=0.7\linewidth]{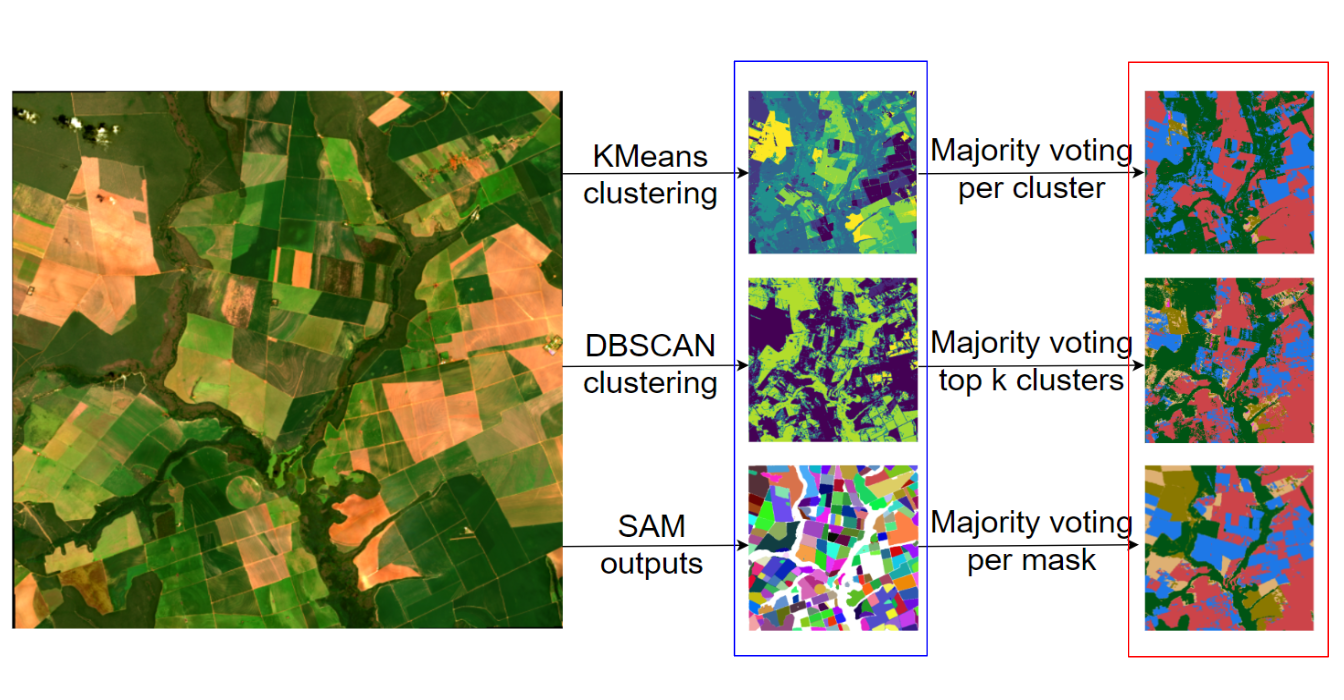}
    \caption{\textbf{Qualitative Comparison}. Comparison of different methods for cleaning up noisy labels. The first column shows the input image. The middle column (blue box) highlights clusters or segments identified by KMeans (\textit{BL1}), DBSCAN (\textit{BL2}), and the proposed approach. The right column (red box) displays the corresponding denoised labels after majority voting within each cluster or segment. The proposed method (third row) using SAM outperforms the traditional clustering methods.}
    \label{fig:result_comparison}
\end{figure*}

\begin{table*}[]
\centering
\caption{\textbf{Performance Comparison of UNet Models on LULC Classification}.
Comparison of accuracy (\textbf{A}), precision (\textbf{P}), and recall (\textbf{R}) for two models, UNet\textsubscript{baseline} and UNet\textsubscript{denoised}, across different land cover classes. The proposed model, UNet\textsubscript{denoised}, demonstrates superior or equal performance in most categories, with bold values indicating the best-performing metrics.}
\label{tab:segment_results}
\begin{tabular}{|c|ccc|ccc|ccc|ccc|ccc|}
\hline
\multirow{2}{*}{\textbf{Model}} & \multicolumn{3}{c|}{\textbf{Cropland}}                                            & \multicolumn{3}{c|}{\textbf{Forest}}                                              & \multicolumn{3}{c|}{\textbf{\begin{tabular}[c]{@{}c@{}}Barren/\\ Built-up\end{tabular}}} & \multicolumn{3}{c|}{\textbf{Waterbody}}                                           & \multicolumn{3}{c|}{\textbf{Pasture}}                                             \\ \cline{2-16} 
                                & \multicolumn{1}{c|}{\textbf{A}}  & \multicolumn{1}{c|}{\textbf{P}}  & \textbf{R}  & \multicolumn{1}{c|}{\textbf{A}}  & \multicolumn{1}{c|}{\textbf{P}}  & \textbf{R}  & \multicolumn{1}{c|}{\textbf{A}}     & \multicolumn{1}{c|}{\textbf{P}}    & \textbf{R}    & \multicolumn{1}{c|}{\textbf{A}}  & \multicolumn{1}{c|}{\textbf{P}}  & \textbf{R}  & \multicolumn{1}{c|}{\textbf{A}}  & \multicolumn{1}{c|}{\textbf{P}}  & \textbf{R}  \\ \hline
UNet\textsubscript{baseline}                  & \multicolumn{1}{c|}{82}          & \multicolumn{1}{c|}{\textbf{91}} & 86          & \multicolumn{1}{c|}{76}          & \multicolumn{1}{c|}{83}          & 75          & \multicolumn{1}{c|}{97}             & \multicolumn{1}{c|}{87}            & \textbf{86}   & \multicolumn{1}{c|}{98}          & \multicolumn{1}{c|}{95}          & 94          & \multicolumn{1}{c|}{82}          & \multicolumn{1}{c|}{81}          & \textbf{87} \\ \hline
UNet\textsubscript{denoised}              & \multicolumn{1}{c|}{\textbf{85}} & \multicolumn{1}{c|}{89}          & \textbf{91} & \multicolumn{1}{c|}{\textbf{79}} & \multicolumn{1}{c|}{\textbf{83}} & \textbf{81} & \multicolumn{1}{c|}{\textbf{97}}    & \multicolumn{1}{c|}{\textbf{91}}   & 83            & \multicolumn{1}{c|}{\textbf{98}} & \multicolumn{1}{c|}{\textbf{95}} & \textbf{96} & \multicolumn{1}{c|}{\textbf{82}} & \multicolumn{1}{c|}{\textbf{83}} & \textbf{87} \\ \hline
\end{tabular}
\end{table*}

\noindent \textbf{C. Impact on Downstream Segmentation Tasks}

The performance of the segmentation model trained on noisy data and denoised data is provided in Table. \ref{tab:segment_results}. The results presented in Table I highlight the effectiveness of the proposed UNet\textsubscript{denoised} model compared to the UNet\textsubscript{baseline}. Across all land cover classes, the UNet\textsubscript{denoised} model consistently outperforms or matches the baseline in terms of accuracy, precision, and recall. Notably, the improvements are most pronounced in the Cropland, Waterbody, and Barren/Built-up classes, where the model trained using denoised labels achieves higher precision and recall, indicating a significant reduction in classification errors. Both models were evaluated on the same held-out test set, ensuring a fair comparison.

\vspace{0.3cm}
\noindent \textbf{D. Comparison with Baselines}

Fig. \ref{fig:result_comparison} shows a comparison of the proposed method with \textit{BL1} and \textit{BL2}. The proposed zero-shot approach of SAM accurately identifies each of the land parcels compared to BL1 and BL2. This further helps reassign the labels resulting in smoother and more precise labels. 

\section{Conclusion}
\label{sec:conclusion}
In this paper, we presented a two-stage approach leveraging foundation models and zero-shot learning to mitigate label noise in LULC datasets. By using the Segment Anything Model (SAM) for precise land parcel segmentation and applying statistical relabeling, our method effectively reduces noise and improves the accuracy of LULC maps. Experimental results demonstrate that this approach surpasses traditional methods, enhancing both denoising and segmentation tasks. Future efforts will focus on extending this approach to utilise and develop EO foundation models.


\bibliographystyle{IEEEbib}
\bibliography{strings,refs}

\begin{thebibliography}{1}

\bibitem{esri}
Krishna Karra et~al.,
\newblock ``Global land use / land cover with sentinel 2 and deep learning,''
\newblock in {\em 2021 IEEE International Geoscience and Remote Sensing Symposium IGARSS}, 2021.

\bibitem{mapbiomas}
Carlos~M. Souza et~al.,
\newblock ``Reconstructing three decades of land use and land cover changes in brazilian biomes with landsat archive and earth engine,''
\newblock {\em Remote Sensing}, 2020.

\bibitem{kmeans}
J.~A. Hartigan and M.~A. Wong,
\newblock ``Algorithm as 136: A k-means clustering algorithm,''
\newblock {\em Journal of the Royal Statistical Society. Series C (Applied Statistics)}, 1979.

\bibitem{dbscan}
Martin Ester et~al.,
\newblock ``A density-based algorithm for discovering clusters in large spatial databases with noise,''
\newblock in {\em Proceedings of the Second International Conference on Knowledge Discovery and Data Mining}. 1996, KDD'96, AAAI Press.

\bibitem{zeroshot}
Yongqin Xian, Christoph~H. Lampert, Bernt Schiele, and Zeynep Akata,
\newblock ``Zero-shot learning—a comprehensive evaluation of the good, the bad and the ugly,''
\newblock {\em IEEE Transactions on Pattern Analysis and Machine Intelligence}, 2019.

\bibitem{sam}
Alexander Kirillov et~al.,
\newblock ``Segment anything,''
\newblock in {\em Proceedings of the IEEE/CVF International Conference on Computer Vision}, 2023.

\end{thebibliography}

\end{document}